# YOLO11-CR: a Lightweight Convolution-and-Attention Framework for Accurate Fatigue Driving Detection

Zhebin Jin[a] and Ligang Dong[a,*]

a. School of Information and Electronic Engineering (Sussex Artificial Intelligence Institute), Hangzhou 310018, China

Email: donglg@zjgsu.edu.cn

## Abstract

Driver fatigue detection is of paramount importance for intelligent transportation systems due to its critical role in mitigating road traffic accidents. While physiological and vehicle dynamics-based methods offer accuracy, they are often intrusive, hardware-dependent, and lack robustness in real-world environments. Vision-based techniques provide a non-intrusive and scalable alternative, but still face challenges such as poor detection of small or occluded objects and limited multi-scale feature modeling. To address these issues, this paper proposes YOLO11-CR, a lightweight and efficient object detection model tailored for real-time fatigue detection. YOLO11-CR introduces two key modules: the Convolution-and-Attention Fusion Module (CAFM), which integrates local CNN features with global Transformer-based context to enhance feature expressiveness; and the Rectangular Calibration Module (RCM), which captures horizontal and vertical contextual information to improve spatial localization, particularly for profile faces and small objects like mobile phones. Experiments on the DSM dataset demonstrated that YOLO11-CR achieves a precision of 87.17%, recall of 83.86%, mAP@50 of 88.09%, and mAP@50-95 of 55.93%, outperforming baseline models significantly. Ablation studies further validate the effectiveness of the CAFM and RCM modules in improving both sensitivity and localization accuracy. These results demonstrate that YOLO11-CR offers a practical and high-performing solution for in-vehicle fatigue monitoring, with strong potential for real-world deployment and future enhancements involving temporal modeling, multi-modal data integration, and embedded optimization.

## 1. Introduction

Fatigue driving has emerged as a critical public safety issue due to the rapid global expansion of vehicle usage [1]. With drivers and vehicles grow annually, fatigue is directly responsible for approximately 20% of all traffic accidents, often resulting in severe injuries or fatalities [2]. Therefore, timely detection and intervention are crucial for mitigating these risks, making the development and deployment of real-time and accurate fatigue detection and alert systems essential strategies for enhancing overall traffic safety and minimizing accident-induced injuries and fatalities.

Current approaches in driver fatigue detection can be broadly categorized into three groups: physiological signal-based methods [3], vehicle dynamics-based approaches [4], and vision-based techniques [5-7]. Physiological signal-based techniques, such as electroencephalography [8-10], provide high accuracy but require invasive measurements [11-13], face computational complexity in signal integration, and are sensitive to environmental and individual variations [14-18]. Vehicle dynamics-based methods, inferring fatigue from abnormal driving behaviors like steering angle fluctuations [19-21], allow non-intrusive monitoring yet lack consistency across diverse driving conditions and vehicle models.

Vision-based methods have gained increasing traction due to their non-intrusive nature, ease of deployment, and compatibility with real-time applications. These methods have evolved from traditional hand-crafted feature methods to advanced deep learning frameworks. Early approaches focused on extracting visual cues such as eye states, yawning, and head poses using manually designed features. Deng et al. [22] developed the DriCare system, which captured blink frequency and yawning from video frames to infer fatigue states, demonstrating the feasibility of vision-based non-intrusive monitoring. Knapik et al. [23] introduced a thermal imaging-based yawn detection method, addressing the limitation of visible-light systems under varying illumination. Saurav et al. [24] and Lima et al. [25] explored eye state recognition by combining convolutional neural networks (CNNs) with support vector machines (SVMs), enabling real-time blink detection. These methods, though effective, relied on pre-defined features and struggled with complex scenarios like partial occlusions or pose variations.

Hybrid models integrating classical machine learning with feature extraction mechanisms also emerged to further improve the effectiveness of vison-based methods. Magan et al. [26] combined CNNs, recurrent neural networks (RNNs), and fuzzy logic to enhance fatigue monitoring, while Younes et al. [27] achieved 92% detection accuracy by fusing RNNs with 3D CNNs, highlighting the value of spatiotemporal modeling. Comparative studies, such as Norah et al. [28], validated MobileNet-V2 as a top performer with 99.6% accuracy on augmented datasets. Zhao et al. [29] integrated MediaPipe Face Mesh with MobileNetV3 and LSTM to achieve 98.4% accuracy on customized data, demonstrating the potential of lightweight architectures.

The functionalities and efficiency of vision-based fatigue detection is further improved with the implementation of end-to-end object detection frameworks, with YOLO models enabling their real-time inference capabilities. In recent years, researchers leveraged YOLO variants to detect fatigue-related scenarios such as closed eyes, head tilts, and mobile phone usage [26,30,31]. Guo et al. proposed a driver fatigue detection method based on YOLOv5, achieving 75.44% mAP on BioID, 80.61% mAP on GI4E, and 43 FPS on GTX 1650 [32]. Wang et al. proposed an improved YOLOv5 model, which achieved a 2.4% increase in mAP compared with the original model, while the FPS only decreased by 8.3[33]. However, with the continuous improvement of model performance requirements, the improvements based on early YOLO versions have gradually shown limitations in terms of adaptability to complex scenarios and multi-dimensional feature fusion, which has driven researchers to conduct deeper explorations based on newer versions of YOLO models.

Notably, recent efforts have focused on integrating advanced attention mechanisms to address these challenges. For instance, Li et al. introduced a Channel-Spatial Attention Module (CSAM) into YOLOv4, enhancing feature representation for small objects by dynamically weighting channel and spatial dimensions [34]. Similarly,

Chen et al. proposed a Multi-scale Feature Attention Network (MFAN) that adaptively aggregates features across different scales, improving detection accuracy for fatigue cues under varying lighting conditions [35]. These studies demonstrated that attention mechanisms can effectively mitigate the limitations of traditional CNN architectures in capturing fine-grained details.

With the release of YOLOv8, studies adopted its transformer-enhanced backbone to capture complex spatial features. Zhang et al. constructed a hybrid model merging YOLOv8 with an LSTM temporal module, improving microsleep detection sensitivity by integrating spatial-temporal context. More recently, the YOLO11 series gained attention for its balance of accuracy and efficiency. Huang et al. [36] developed LW-YOLO11, a lightweight variant that reduced computational load while maintaining precision for profile face and phone detection. Deng et al. [37] integrated multi-scale attention into YOLOv6 to improve sensitivity to subtle facial fatigue indicators under varying lighting. These advancements showcased a trend toward multi-object, multi-scale detection frameworks. YOLO models excelled in balancing high mAP with real-time processing, making them suitable for in-vehicle deployment. Key innovations included attention mechanisms, multi-scale feature aggregation, and anchor-free detection heads, which enhanced the ability to recognize diverse fatigue cues from facial expressions to secondary task distractions.

Nevertheless, vision-based methods have faced critical hurdles. Robustness under complex conditions, such as partial occlusions by hands or sunglasses, side-view facial poses, and low-light environments, which have remained challenging; for example, standard models struggled to detect closed eyes when obstructed by sunglasses. Recognizing simultaneous behaviors like phone usage and yawning required advanced spatial-temporal reasoning, which many models lacked [22]. Models trained on limited datasets often failed to adapt to diverse driving environments, camera angles, or driver demographics [38-40]. While lightweight models emerged, real-time performance on low-power embedded systems remained a bottleneck for widespread adoption. Future research should focus on enhancing feature representation for occluded objects, integrating multi-modal sensing, such as thermal and visible light, and developing domain-adaptive models. Temporal modeling and cross-scenario validation via large-scale field trials will also be essential to bridge the gap between laboratory accuracy and real-world reliability.

To address these limitations, including poor robustness under occlusion, limited multi-scale feature modeling, and high computational costs. This paper proposes YOLO11-CR, a lightweight and high-performance detection model. The main contributions of this paper can be summarized as:

1) designing the Convolution and Attention Fusion Module (CAFM) to replace the attention layer in the C2PSA module, forming an enhanced C2PSA_CAFM structure. This module integrates CNNs and Transformers via local and global branches for extracting respective features, with the final output from summing these streams to model local-global representations and enhance feature expressiveness and contextual understanding.
2) introducing the Rectangular Calibration Module (RCM) to replace conventional 3 × 3 and 1 × 1 convolutions in specific feature extraction/fusion stages of YOLO11n. RCM enhances spatial feature modeling in YOLO11 by capturing horizontal-vertical global context, enabling more accurate multi-scale object localization/recognition and improving overall network detection performance.
3) fine-tuning and testing YOLO11-CR for fatigue detection scenarios, focusing on three key target categories: normal face, profile face, and mobile phone.

The remainder of the paper is organized as follows: Section 2 provides a detailed introduction to proposed YOLO11-CR, the structure of the CAFM, and the structure of the RCM. Section 3 introduces the experimental setup, including an overview of the dataset, training hyperparameters, and evaluation metrics. Section 4 conducts ablation studies on the CAFM and RCM modules in YOLO11-CR, followed by a comprehensive analysis of performance parameters. Finally, this paper is concluded in Section 5.

## 2. Framework

To build an effective fatigue driving detection system, this paper proposes YOLO11-CR, an enhanced one-stage object detection framework based on YOLO11, to address challenges of small-scale objects, partial occlusions, and non-frontal facial orientations in driver behavior analysis. As shown in Fig. 1, YOLO11-CR incorporates two novel architectural modules: the CAFM and RCM, which collaboratively enhance multi-scale representation learning, spatial feature alignment, and detection accuracy under complex scenarios. Adopting a typical encoder-decoder design, the network comprises a YOLO11n backbone for hierarchical feature extraction, a feature fusion neck, and a multi-scale detection head. The detection targets are defined as frontal face, profile face, and mobile phone, which serve as critical semantic cues in fatigue detection systems. The backbone generates P3 – P5 level features upsampled by factors of 8, 16, and 32, which are augmented with attention-enhanced representations before being forwarded to the detection head for bounding box regression and class probability prediction.

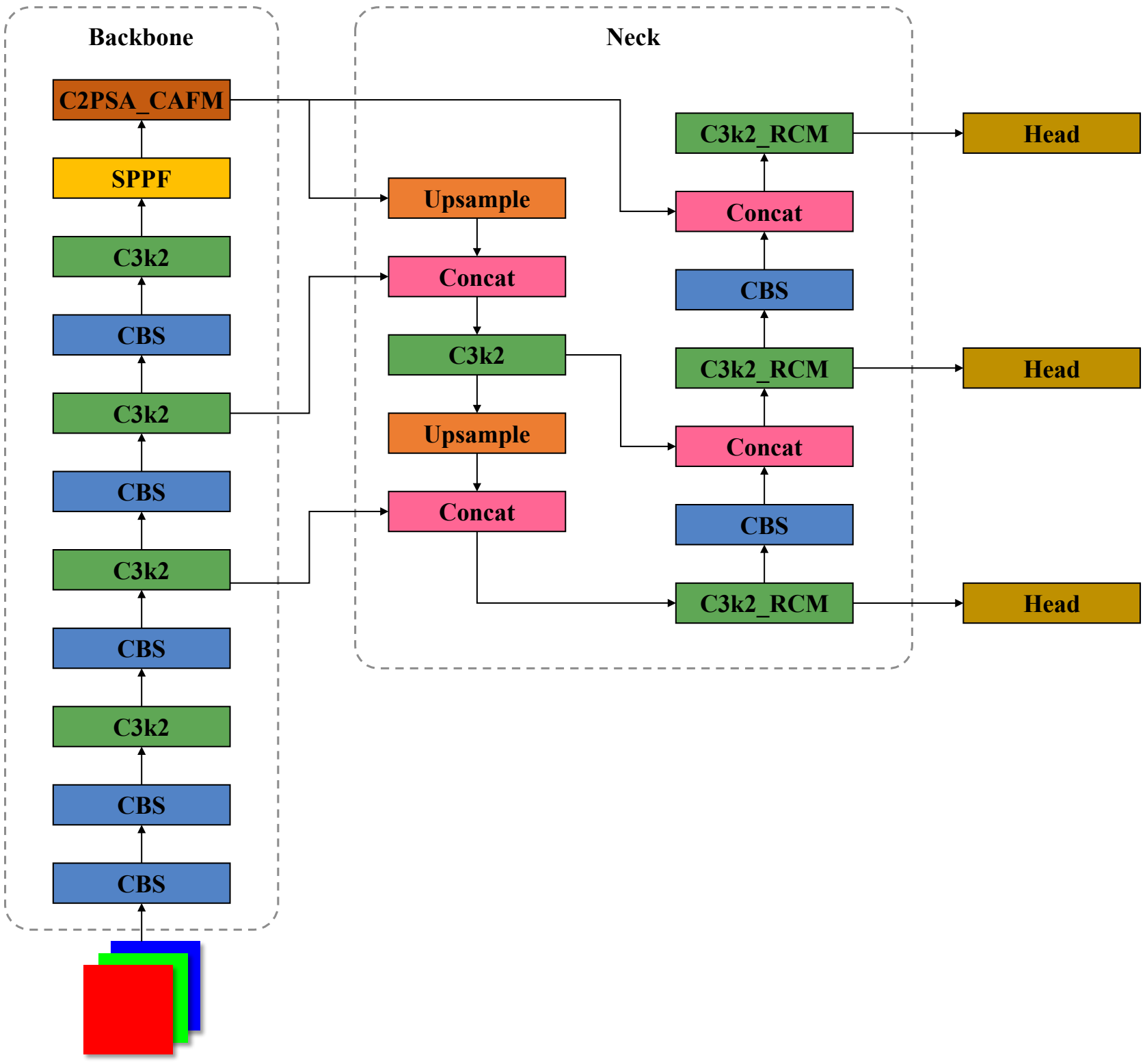

**Fig. 1** Framework of the Proposed YOLO11-CR.

## 2.1 Convolution-and-Attention Fusion Module

To addresses the fundamental challenge of simultaneously capturing local fine-grained features and global contextual relationships in a lightweight and efficient manner. Inspired by the complementary strengths of convolutional operations and self-attention mechanisms, CAFM [41] is introduced in this section to solve the problem of complex detection scenarios, such as small object detection, occluded object recognition, and fatigue feature extraction.

As illustrated in Fig. 2, the CAFM consists of two functional branches, the local branch, designed to capture fine-grained spatial patterns crucial for detecting small-scale targets and maintaining boundary precision, extracts spatial details through convolutional operations, the global branch addresses the limited receptive field of convolutions by introducing a lightweight self-attention mechanism to model long-range spatial dependencies, which are essential for understanding occluded or distributed features.

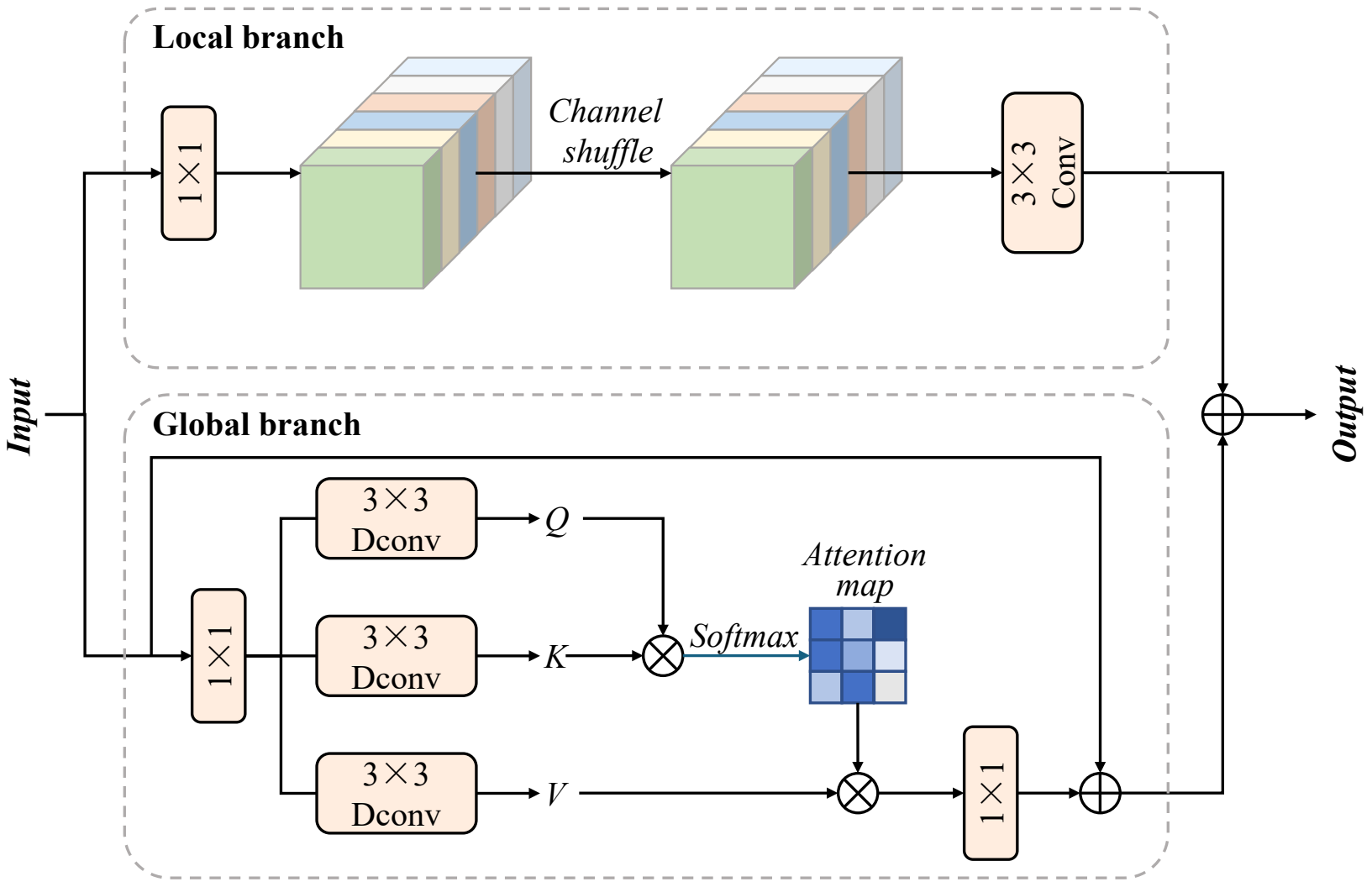

**Fig. 2** The Structure of CAFM.

Given the input feature map $Y$, the local branch sequentially applies a $1 \times 1$ convolution $W_{1\times1}$ to adjust channel dimensions and enhance feature interaction, then a channel shuffle operation $CS$ to improve inter-channel information flow, and a $3 \times 3$ depthwise convolution $W_{3\times3\times3}$ to further extract spatial localized features, with the entire processing flow can be calculated as:

$$F_{conv} = W_{3\times3\times3}\left(CS\left(W_{1\times1}(Y)\right)\right) \tag{1}$$

While the global branch first generates Query (Q), Key (K), and Value (V) matrices via $1 \times 1$ convolution and depthwise $3 \times 3$ convolution, then reduces computational cost by computing attention across feature channels instead of the full spatial domain, and finally applies scaled dot-product attention formulated as:

$$Attention(Q, K, V) = V \cdot Softmax\left(\frac{K^T Q}{\alpha}\right) \quad (2)$$

Where $\alpha$ is a learnable scaling parameter controlling the softmax sharpness. And then, the attention-enhanced output can be calculated:

$$F_{attn} = W_{1\times1}(Attention(Q, K, V) + Y) \quad (3)$$

Finally, the fused output $F_{out}$ can be calculated:

$$F_{out} = F_{conv} + F_{attn} \quad (4)$$

## 2.2 Rectangular Calibration Module

Although traditional convolutional operations and standard attention mechanisms are effective, they often struggle to precisely model elongated, axially aligned, and partially occluded structures commonly found in real-world scenes, such as profile faces, hand-held objects, or fatigue-related gestures. To address these challenges, the RCM [42] is incorporated into the network, as shown in Fig. 3, RCM comprises four key components: axial global context aggregation, shape self-calibration reconstruction, local-global feature fusion, and residual refinement.

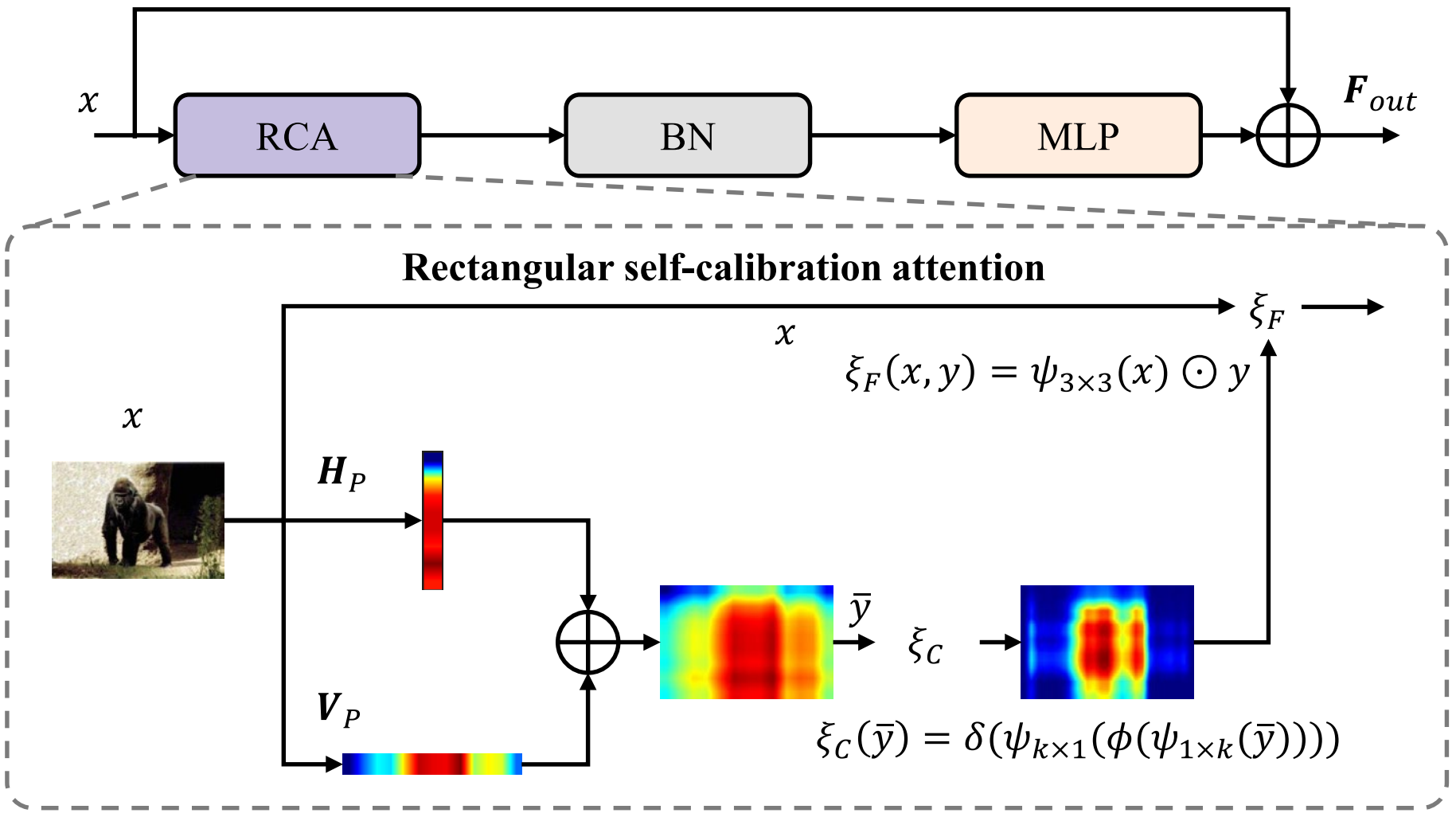

**Fig. 3** The Structure of RCM.

To model rectangular regions, RCM decomposes 2D global attention into horizontal pooling (HP) that averages feature responses along each row to capture horizontal context and vertical pooling (VP) that averages responses along each column to capture vertical context, with these two axial global contexts broadcast-added to form an initial coarse rectangular attention map as:

$$RectAttention(x) = HP(x) \oplus VP(x) \tag{5}$$

Where $x$ is the input feature map, and $\oplus$ denotes element-wise broadcast addition.

Since the initial rectangular attention map may remain coarse or fail to match actual object contours, RCM incorporates a shape self-calibration function that dynamically adjusts the attention shape through a process involving large-kernel strip convolutions, followed by Batch Normalization and ReLU activation function, the operation is mathematically defined as:

$$\xi C(x) = \delta\left(\psi_{k\times 1}\left(\phi(\psi_{1\times k}(\bar{y}))\right)\right) \tag{6}$$

Where $\psi$ denotes strip convolutions, $\phi$ denotes BatchNorm and ReLU, and $\delta$ is the Sigmoid activation to constrain outputs between 0 and 1.

To enrich the recalibrated attention with local spatial details, the input feature $x$ is first processed by a 3×3 depthwise convolution, and then the calibrated attention map is element-wise multiplied with the locally refined feature map. The local-global feature fusion can be calculated as follows:

$$\xi F(x, y) = \psi_{3\times 3} \odot y \tag{7}$$

Where $\odot$ denotes Hadamard multiplication. Through local-global feature fusion, the final feature can ensure the retention of crucial local texture details while fully leveraging the global context information, thereby providing strong support for precise detection.

Inspired by residual learning strategies, the output of RCM is processed via a lightweight MLP and batch normalization layer, with a residual connection added:

$$F_{out} = \rho\left(\xi F\left(x, \xi C\left(HP(x) \oplus VP(x)\right)\right)\right) + x \tag{8}$$

Where $\rho$ represents a BatchNorm followed by a MLP transformation, enabling enhanced feature reuse and stabilized gradient flow via the residual connection.

**2.3 Multi-Scale Detection Head**

The multi-scale detection head operates at three resolutions (P3, P4, P5), each embedded with an RCM block to refine semantic representations prior to prediction, where the final Detect module predicts class probabilities and bounding boxes for three categories across scales, and by incorporating CAFM at the backbone tail and RCM across detection scales, YOLO11-CR effectively improves the detection of normal faces via fused attention for global facial structure capture, profile faces via rectangular kernels for elongated contour localization, and mobile phones via multi-scale context and directional attention for background noise and clutter suppression.

## 3. Experiments Setup

To validate the effectiveness of the proposed YOLO11-CR model for fatigue detection, a comprehensive set of experiments was designed and conducted. This section details the experimental setup, including the datasets selection, evaluation metrics, and implementation details.

### 3.1 Experimental Dataset

To evaluate the effectiveness of the proposed YOLO11-CR model in detecting fatigue-related behaviors, this study employs the Driver State Monitoring (DSM) Dataset, a comprehensive benchmark specifically curated for driver fatigue and distraction detection. The DSM dataset is derived from the publicly available DMD dataset introduced by Ortega et al., which is one of the most widely adopted multi-modal driver monitoring datasets in the field.

The DSM dataset contains over 180,000 RGB images extracted from in-vehicle video recordings covering diverse environments such as urban roads, highways, and nighttime driving. The data was collected using high-definition cameras placed on the dashboard and ceiling, ensuring coverage of both frontal and profile views. As shown in Fig. 4, several typical samples from the dataset illustrate the diversity of captured behaviors, including frontal facial expressions and profile postures. Each image is annotated with behavior-specific labels including “normal driving”, “eye closure”, “yawning”, “head dropping”, “phone usage,” and “looking sideways”, and the detail label distribution is detailed in Table 1. The dataset includes annotations for bounding boxes, facial landmarks, and occlusion levels, enabling robust evaluation of detection models under varied visual conditions. Furthermore, the dataset incorporates various demographic features such as different age groups, genders, and accessories (e.g., sunglasses, hats, masks), making it ideal for fatigue detection tasks that require generalization across drivers. For the purpose of model training and evaluation, the dataset was partitioned into training, validation and test at a 7:2:1 ratio.

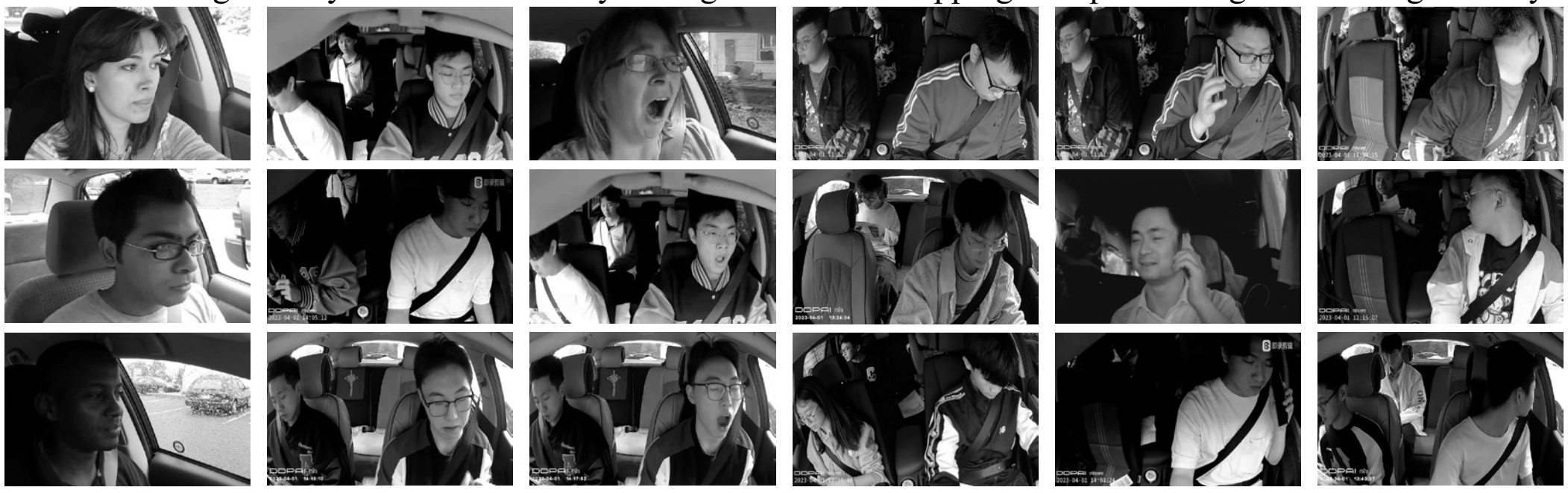

**Fig. 4** Representative Images of the DSM Dataset.

**Table 1** The Label Distribution of DSM dataset.

| Behavior Type | Number of Samples | Percentages (%) |
|---|---|---|
| Normal Driving | 45,000 | 25.0 |
| Eye Closure | 36,000 | 20.0 |
| Yawning | 30,000 | 16.7 |
| Head Dropping | 27,000 | 15.0 |
| Phone Usage | 21,000 | 11.7 |
| Looking Sideways | 21,000 | 11.7 |
| Total | 180,000 | 100.0 |

In addition to frame-wise annotations, DSM dataset supports temporal analysis, allowing the application of sequential models to detect gradual-onset fatigue symptoms such as blinking frequency reduction or microsleeps. As highlighted by Ortega et al., the multimodal structure and real-world complexity make it an ideal benchmark for testing fatigue detection systems in safety-critical automotive applications.

### 3.2 Implementation Details

All experiments were conducted using a high-performance computing environment equipped with an AMD Ryzen 9 5950X CPU, a NVIDIA GeForce RTX 3090 GPU, and 64 GB of DDR5 RAM. The operating system was Windows 11, and the implementation was carried out using Python 3.11.2 and PyTorch 2.5.1 with CUDA 12.4 support. This configuration ensured sufficient computational resources to handle large-scale training tasks efficiently.

During training, the batch size was set to 64, and the Stochastic Gradient Descent (SGD) optimizer was utilized with an initial learning rate of 0.001. The learning rate was adjusted dynamically during training using a cosine annealing scheduler. The momentum coefficient for SGD was set to 0.937 to facilitate stable convergence. The model was trained for a total of 100 epochs. Early stopping mechanisms were not employed, allowing the model to complete the full training schedule and explore the optimization landscape thoroughly.

To enhance the generalization ability and robustness, data augmentation techniques were applied during training. Specifically, Mosaic augmentation and random horizontal flipping were incorporated to increase the diversity of training samples and to simulate various real-world driving conditions.

### 3.3 Evaluation Metrics

To evaluate the performance of the proposed YOLO11-CR model, three primary evaluation metrics in object detection, e.g., mean Average Precision at an IoU threshold of 0.5 (mAP@50), mean Average Precision over a range of IoU thresholds from 0.5 to 0.95 (mAP@50 - 95), Precision, and Recall. These metrics can be expressed as:

$$AP = \int_0^1 P(R)dR \tag{9}$$

$$mAP = \frac{\sum_{i=1}^{K} AP_i}{K} \times 100\% \tag{10}$$

$$Precision = \frac{TP}{TP + FP} \times 100\% \tag{11}$$

$$Recall = \frac{TP}{TP + FN} \times 100\% \tag{12}$$

Where *TP* represents true positives that are correctly predicted, *FP* represents false positives that are incorrectly predicted as positives, and *FN* represents false negatives that are incorrectly predicted as negatives. *K* represents the number of classes.

## 4. Experiment Results and Discussions

To comprehensively evaluate the effectiveness of the proposed YOLO11-CR model, a series of comparative experiments were conducted against several baseline detection models. This section presents a detailed analysis of the results, examining the impact of individual module enhancements through ablation studies, per-class detection accuracy, confusion matrix insights, precision-recall characteristics, and comparison with state-of-the-art (SOTA) models.

### 4.1 Ablation Experiments

To evaluate the contributions of the CAFM and the RCM to the overall performance improvements, an ablation study was conducted. The results are shown in Table 2.

**Table 2** Ablation Experiments Results on DSM dataset Testing Set.

| Model Variant | Precision | Recall | mAP@50 | mAP@50-95 |
|---|---|---|---|---|
| YOLO11n | 0.8648 | 0.7622 | 0.8228 | 0.5337 |
| YOLO11n+RCM | **0.8810** | 0.7748 | 0.8494 | 0.5358 |
| YOLO11n+CAFM | 0.8466 | 0.7942 | 0.8416 | 0.5363 |
| YOLO11-CR | 0.8717 | **0.8386** | **0.8809** | **0.5593** |

Adding the RCM module to the baseline model led to an increase in Precision (+1.62%) and mAP@50 (+2.66%) with a minor improvement in Recall (+1.23%). This suggests that the RCM primarily enhances the localization ability of the model, enabling more accurate bounding box regression. The rectangular calibration effectively aligns the predicted boxes to the natural aspect ratios of human faces and mobile devices, thus improving the precision of detection without compromising recall.

Incorporating the CAFM module resulted in a substantial Recall improvement (+3.20%), while the Precision slightly decreased (−1.82%) compared to the baseline. This indicates that the CAFM strengthens the feature extraction capability, allowing it to capture more challenging or partially occluded objects. However, the slight drop in Precision implies that the feature fusion process may introduce a few more false positives, especially when distinguishing small handheld devices from background noise.

When both CAFM and RCM were integrated, the model achieved the highest Precision, Recall, mAP@50, and mAP@50-95 simultaneously. The combined model did not merely aggregate the individual improvements but further enhanced overall robustness, suggesting a strong complementary relationship between CAFM and RCM. While CAFM improves detection sensitivity, RCM refines spatial calibration, leading to significant gains in both detection sensitivity and localization precision.

Given that the mobile phone category typically involves small-scale objects, the combined improvements in recall and precision indicate that YOLO11-CR is particularly effective at detecting small, occluded, and non-frontal targets. This validates the intuition that advanced multi-scale feature fusion and adaptive spatial calibration are critical for improving driver monitoring systems under real-world conditions.

Overall, the ablation study confirms that both modules contribute uniquely and synergistically to the enhanced performance of the YOLO11-CR model.

**4.2 Per-Class Performance Analysis**

To further investigate the effectiveness across different fatigue-related classes, per-class evaluation metrics were computed, as shown in Table 3, it can be observed that the model achieves outstanding detection performance on the normal face class, with a Precision of 93.0%, Recall of 95.9%, and an exceptionally high mAP@50 of 98.6%. This demonstrates strong capability in capturing clear, frontal facial features even under diverse environmental conditions.

**Table 3** Per-Class Performance Results of YOLO11-CR.

| Class | Precision | Recall | mAP@50 | mAP@50-95 |
|---|---|---|---|---|
| Normal Face | 0.930 | 0.959 | 0.986 | 0.821 |
| Profile Face | 0.909 | 0.754 | 0.891 | 0.483 |
| Mobile Phone | 0.758 | 0.659 | 0.716 | 0.366 |

The detection performance for the profile face class is slightly lower, particularly in Recall (75.4%), suggesting that side poses are more challenging to detect due to less distinguishable features and higher variance in facial appearances. Nevertheless, the achieved mAP@50 of 89.1% remains satisfactory for practical applications.

The mobile phone class exhibits the lowest detection metrics among the three categories, with a Precision of 75.8%, Recall of 65.9%, and a mAP@50-95 of only 36.6%. This is primarily because mobile phones appear smaller in size and are often partially occluded by hands or steering wheels, making them more difficult to detect. Despite these challenges, the YOLO11-CR model still maintains reasonable performance, thanks to the enhanced feature extraction capabilities introduced by the CAFM and the adaptive spatial adjustment enabled by the RCM.

### 4.3 Confusion Matrix Analysis

To gain deeper insights into the model performance, the normalized confusion matrices for the four evaluated models were analyzed, as shown in Fig. 5. Each confusion matrix presents the prediction accuracy across the four classes: profile face (pface), normal face (nface), mobile phone, and background.

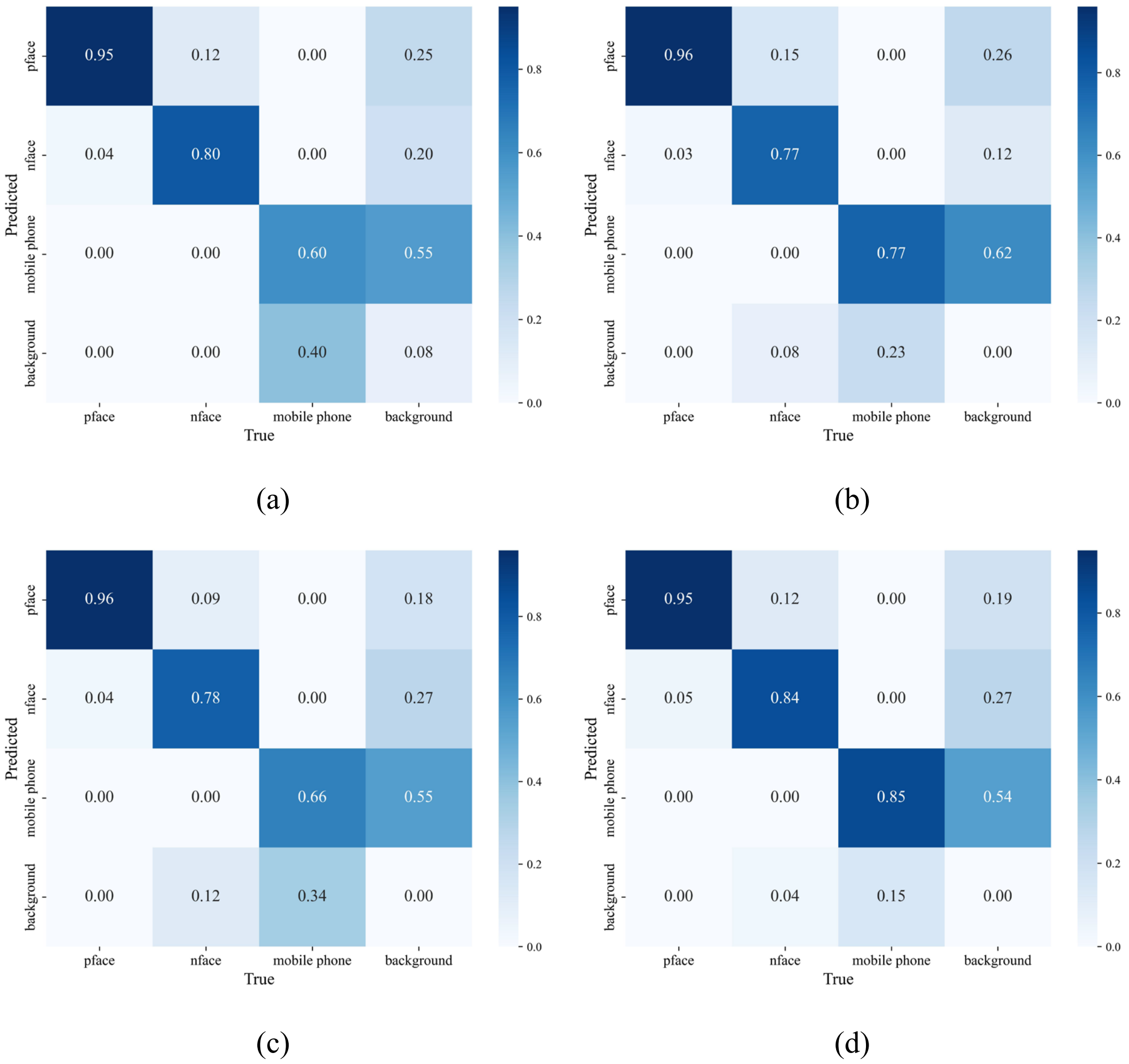

**Fig. 5** The Group of Confusion Matrix: (a) YOLOv8, (b) YOLOv10n, (c) YOLO11n, and (d) YOLO11-CR.

Across all models, the pface category consistently demonstrated the highest classification accuracy, with YOLO11-CR achieving a correct prediction rate of 96%, outperforming YOLOv8, YOLOv10n, and YOLO11n. This indicates that frontal faces, with their distinct and easily recognizable features, are relatively easier to detect.

For the nface category, YOLO11-CR exhibited a correct classification rate of 84%, higher than that of YOLOv10n (77%) and YOLO11n (78%), and slightly surpassing YOLOv8 (80%). This result highlights the improved robustness of YOLO11-CR in handling partial facial views, likely due to the enhanced global-local feature extraction enabled by the CAFM module.

Detection performance for the mobile phone category revealed significant differences among the models. YOLO11-CR achieved a correct classification rate of 85%, substantially outperforming YOLOv8 (60%), YOLOv10n (77%), and YOLO11n (66%). This underscores the model enhanced ability to detect small and occluded objects, a direct benefit of the RCM module's adaptive spatial calibration.

The background category showed relatively lower classification consistency across all models, which is expected due to the diverse and complex nature of real-world driving environments. However, YOLO11-CR maintained a reasonably high true negative rate, minimizing false positives for fatigue-unrelated regions.

Overall, the confusion matrix analysis further confirms that YOLO11-CR significantly improves both the classification sensitivity (higher Recall for challenging classes like side face and mobile phone) and classification specificity (lower false positives in background regions), thereby achieving a balanced and robust performance suitable for real-world driver monitoring applications.

### 4.4 Precision-Recall Curve Analysis

To further evaluate the detection performance across different models, Precision-Recall (PR) curves for each model were analyzed, as illustrated in Fig. 6. The PR curves depict the relationship between Precision and Recall across various detection thresholds, offering a comprehensive view of the robustness of the model in balancing sensitivity and specificity.

The YOLO11-CR model achieved the most favorable PR curve trajectory among the compared models. As shown in Fig. 6(d), YOLO11-CR maintained a higher precision across the entire range of recall values compared to YOLOv8, YOLOv10n, and YOLO11n. Specifically, even at high recall levels (e.g., Recall > 0.8), YOLO11-CR sustained a precision above 80%, demonstrating its robustness in detecting fatigue-related behaviors without significant performance degradation. In contrast, YOLOv8 and YOLOv10n exhibited a steeper decline in precision as recall increased, particularly in the mobile phone category, where object size and occlusion complexity posed greater challenges. YOLOv8, despite achieving relatively high recall, showed a substantial drop in precision beyond Recall = 0.6, indicating a higher rate of false positives under permissive thresholds.

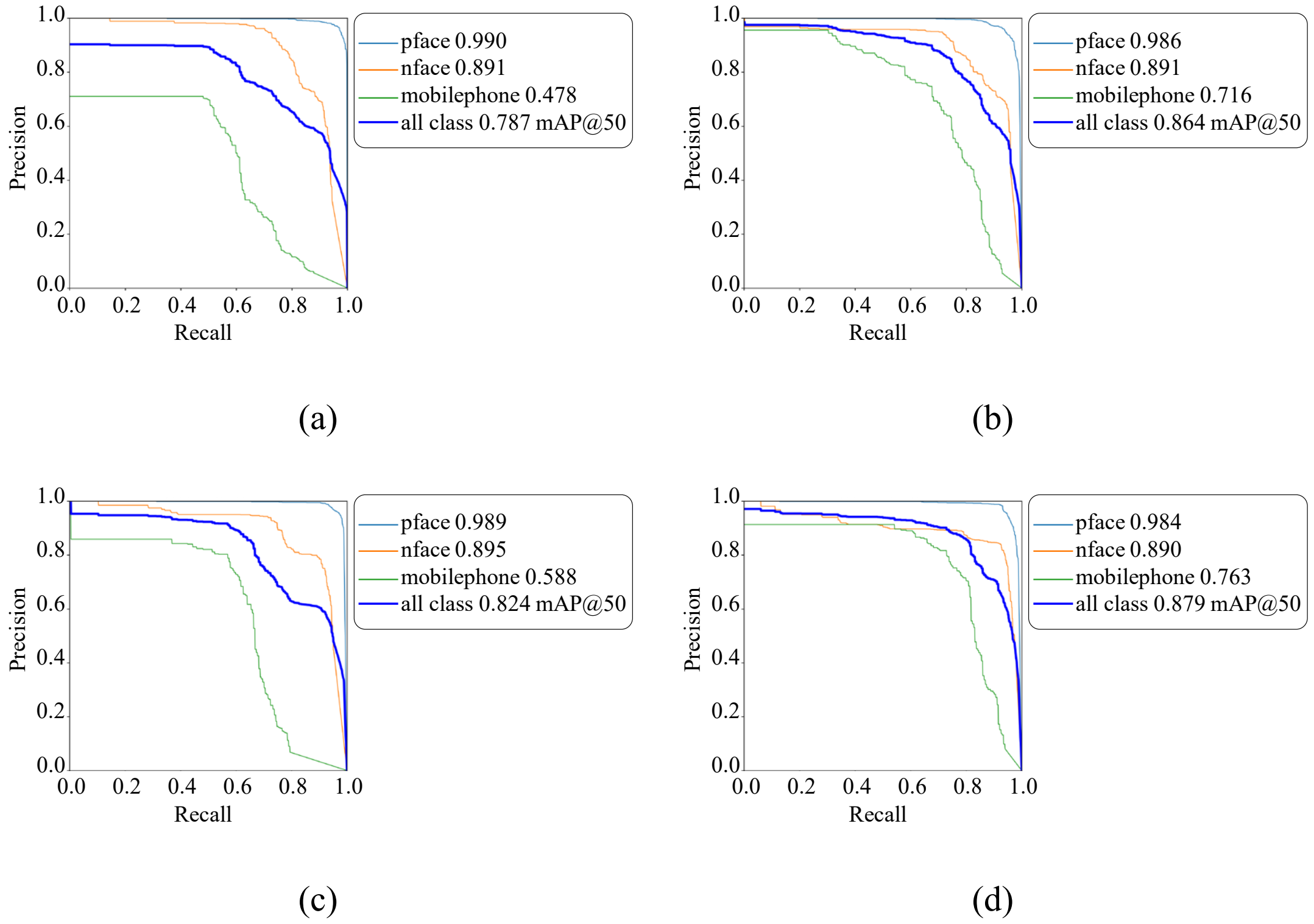

**Fig. 6** The Group of PR Curves: (a) YOLOv8, (b) YOLOv10n, (c) YOLO11n, and (d) YOLO11-CR.

YOLO11-CR, YOLO11n, YOLOv10n, and YOLOv8 achieved mAP@50 of 87.9%, 82.4%, 86.4%, and78.7% respectively, further supporting these observations based on PR curves. Notably, the performance gap between YOLO11-CR and YOLOv8 is especially pronounced in the mobile phone category, where YOLO11-CR achieved a mobile phone mAP of 76.3% compared to only 47.8% for YOLOv8. This highlights the efficacy of the CAFM and RCM in enhancing small object detection.

Overall, the PR curve analysis corroborates the superior precision-recall balance and stability of the YOLO11-CR model, confirming its effectiveness in achieving both high sensitivity and reliability in fatigue detection tasks.

### 4.5 Comparison with SOTA

Finally, to evaluate the effectiveness of the proposed YOLO11-CR model, comparative experiments were conducted against several SOTA models, including YOLOv8, YOLOv10n, and YOLO11n. Table 4 summarized the performance metrics of these models in terms of Precision, Recall, mAP@50, mAP@50-95, parameter, GLOPs, and FPS on the DSM dataset testing subset. Although the FPS metric of YOLO11-CR is not the best, considering that all other performance metrics are optimal, it achieves the ideal balance between efficiency and accuracy.

**Table 4** Performance Comparison with SOTA models on DSM dataset Testing Set

| Model | Precision | Recall | mAP@50 | mAP@50-95 | M params | GLOPs | FPS |
|---|---|---|---|---|---|---|---|
| YOLOv8 | 0.740 | <u>0.796</u> | 0.769 | 0.499 | 3.157 | 4.43 | **184.0** |
| YOLOv10n | 0.856 | 0.761 | <u>0.842</u> | <u>0.543</u> | 2.775 | 4.37 | 111.1 |
| YOLO11n | <u>0.865</u> | 0.762 | 0.823 | 0.534 | **2.624** | <u>3.31</u> | <u>129.6</u> |
| YOLO11-CR | **0.872** | **0.839** | **0.881** | **0.559** | <u>2.691</u> | **3.28** | 106.4 |

Fig. 7 presents the fatigue detection results of different YOLO - series models on driving - scenario images. Each row corresponds to a set of test samples from driving scenarios, covering complex working conditions such as in - vehicle faces and hand - held objects. YOLOv8, YOLOv10n, and YOLOv11n all inevitably suffer from missed detection or false detection in hand - held object detection. In contrast, the YOLO11 - CR model proposed in this paper effectively addresses the issues of missed detection and false detection for long - strip - shaped objects like mobile phones, achieving superior detection accuracy.

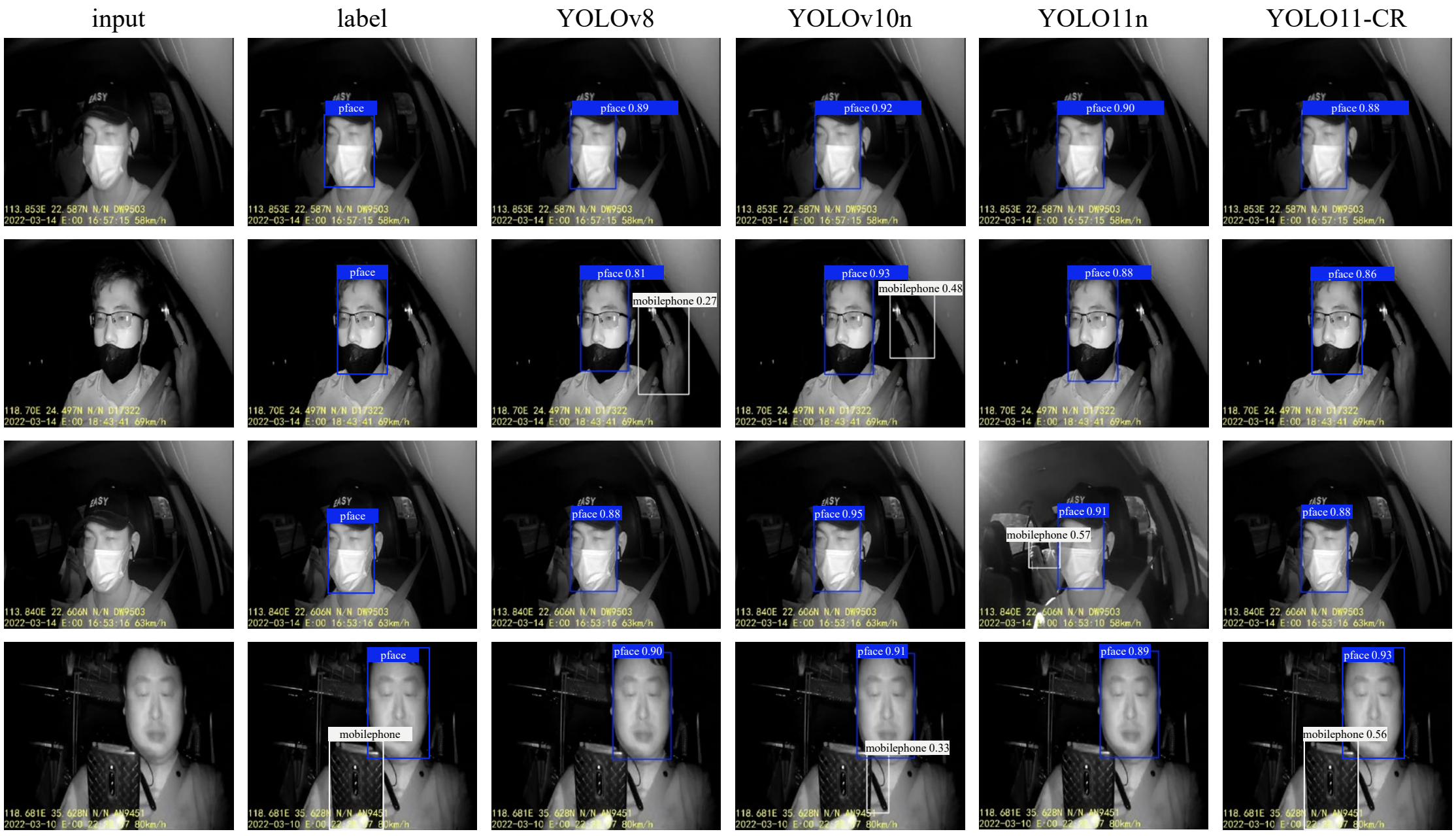

**Fig. 7** Comparison of Driver Fatigue Detection Against State-of-the-Art Models on Various Driving Scenarios.

## 5. Conclusion

This paper presented the design and optimization of a lightweight, high-accuracy fatigue driving detection system based on an improved YOLO11 model, referred to as YOLO11-CR. By integrating the CAFM and the RCM into the baseline YOLO11 architecture, the proposed model significantly enhanced feature extraction capabilities and spatial localization accuracy, particularly for small-scale and occluded objects.

Extensive experimental evaluations on the DSM dataset demonstrated that YOLO11-CR consistently outperformed baseline models across key performance metrics, achieving Precision of 87.17%, Recall of 83.86%, mAP@50 of 88.09%, and mAP@50-95 of 55.93%. Per-class performance analyses, confusion matrix evaluations, and PR curve comparisons further validated the robustness and reliability of the proposed system in real-world fatigue detection scenarios. The ablation study confirmed that the CAFM and RCM modules provide complementary enhancements, with CAFM improving detection sensitivity and RCM refining spatial alignment. Their combined integration resulted in significant synergistic performance gains.

Overall, YOLO11-CR offers a practical, efficient, and robust solution for real-time fatigue monitoring, with strong potential for deployment in intelligent in-vehicle safety systems.